# Improved YOLOv7x-Based Defect Detection Algorithm for Power Equipment

Jin Hou ， Hao Tang

School of Information Science and Technology, Southwest Jiaotong University, Chengdu, Sichuan 611756, China; th520625@my.swjtu.edu.cn

**Abstract:** The normal operation of power equipment plays a critical role in the power system, making anomaly detection for power equipment highly significant. This paper proposes an improved YOLOv7x-based anomaly detection algorithm for power equipment. First, the ACmix convolutional mixed attention mechanism module is introduced to effectively suppress background noise and irrelevant features, thereby enhancing the network's feature extraction capability. Second, the Biformer attention mechanism is added to the network to strengthen the focus on key features, improving the network's ability to flexibly recognize feature images. Finally, to more comprehensively evaluate the relationship between predicted and ground truth bounding boxes, the original loss function is replaced with the MPDIoU function, addressing the issue of mismatched predicted bounding boxes. The improved algorithm enhances detection accuracy, achieving a mAP@0.5/% of 93.5% for all target categories, a precision of 97.1%, and a recall of 97%.

## 1. Introduction

With the rapid development of China's power industry, the coverage area for power equipment installation has been continuously expanding. Among these components, insulators and dials are critical elements. Several incidents have occurred in power grids where equipment failures were caused by the malfunctioning of insulators and dials, highlighting the importance of regularly inspecting images obtained from intelligent inspections [1]. With the advancement of deep learning in image detection, object detection algorithms are increasingly applied in the defect detection of power equipment.

Object detection models are mainly categorized into one-stage models, such as SSD [2], RetinaNet [3], and the YOLO series, and two-stage models, such as Faster R-CNN [4], SPPNet [5], and Mask R-CNN [6]. One-stage models are faster and more suitable for real-time defect detection in power equipment. However, these algorithms often struggle with accuracy when detecting defects in power equipment with complex backgrounds. Since intelligent inspections of power equipment require accurate recognition of various types of defects in complex backgrounds, further optimization of detection algorithms is necessary.

In this regard, numerous studies have been conducted. For example, Ma Qiaohui et al. [7] proposed a fault detection algorithm based on skirt morphological features to address the difficulty of insulator fault detection in power equipment. While this method provides a foundation for the intelligent detection of insulators, it still falls short in evaluating the integrity of insulators. TAO et al. [9] proposed a cascaded convolutional network structure for locating and detecting defects in insulators within power equipment. However, this cascaded network can only detect self-explosion defects, limiting the algorithm's robustness. ZHAO et al. [10] improved the anchor generation and non-maximum suppression (NMS) methods in the Region Proposal Network (RPN) of the Faster R-CNN deep detection model to enhance detection accuracy, but the computational load of this algorithm is high. Song Zhiwei et al. [11] introduced an improved YOLOv7 defect detection method that combines global attention mechanisms to enhance salient features. Although it achieved some success in detecting insulator breakage defects, it still falls short of fully meeting the requirements of intelligent inspections of power equipment. Zhao Wenqing et al. [12] addressed the problem of locating and identifying insulators in images with Faster R-CNN by proposing a method that combines the CBAM attention mechanism. While this improved detection accuracy, it remains challenging to accurately locate defects in complex backgrounds.

Although the aforementioned algorithms have achieved improvements to some extent, they still exhibit limitations in handling complex scenarios and detecting diverse types of power equipment. Most current defect detection algorithms focus on insulators, primarily targeting insulator breakage defects, with limited attention paid to flashover defects or other common defects in power equipment inspections. To meet the requirements of intelligent inspection systems for power equipment, enhance the accuracy of anomaly detection, and reduce false detection and missed detection rates, this paper proposes several innovations.

First, the ACmix convolution and self-attention mixed module is integrated into the SPPCSPC structure, enabling the network to better extract features of various targets. Additionally, convolutional layers are pruned to reduce parameters and computational complexity, and the parallel pooling layers are modified into serial pooling layers. By fusing feature maps from serial pooling layers, the receptive field is expanded while computational speed is improved. The improved SPPCSPC structure enhances the network's ability to recognize anomalous target features of different sizes.

Second, the Biformer attention mechanism is incorporated into the Efficient Aggregation Network (EAGNet) to strengthen the network's focus on key features, improving its ability to flexibly recognize feature images.

Finally, the original loss function is replaced with the MPDIoU function to provide a more comprehensive evaluation of the relationship between predicted and ground truth bounding boxes, thereby improving the model's detection performance.

## 2. Related Technologies

The Materials and Methods should be described with sufficient details to allow others to replicate and build on the published results. Please note that the publication of your manuscript implicates that you must make all materials, data, computer code, and protocols associated with the publication available to readers. Please disclose at the submission stage any restrictions on the availability of materials or information. New methods and protocols should be described in detail while well-established methods can be briefly described and appropriately cited.

Research manuscripts reporting large datasets that are deposited in a publicly available database should specify where the data have been deposited and provide the relevant accession numbers. If the accession numbers have not yet been obtained at the time of submission, please state that they will be provided during review. They must be provided prior to publication.

Interventionary studies involving animals or humans, and other studies that require ethical approval, must list the authority that provided approval and the corresponding ethical approval code.

*2.1 YOLOv7 Network Model*

The backbone of the YOLOv7 model consists primarily of CBS convolutional layers, an efficient aggregation network, the MP-1 module, and the SPPCSPC module. The CBS convolutional layers are used for initial feature extraction from images. The efficient aggregation network enriches feature representation by controlling connection paths of varying lengths. The MP-1 module extracts features through two branches—upper and lower—using max pooling and convolution operations, and then fuses the features via a Concat operation, enabling the network to extract more effective information. The SPPCSPC module integrates features with different receptive fields, avoiding redundant feature extraction from images and enhancing the expressiveness of the feature maps. The neck of the model employs a Path Aggregation Feature Pyramid Network (PAFPN), which extracts semantic features in a top-down manner and combines them with precise localization information.

## 2.2 ACmix Convolutional Mixed Self-Attention Mechanism Module

The ACmix convolutional mixed self-attention mechanism module [17] combines convolution and self-attention to capture both local and global information, improving feature representation capabilities. The module operates in two stages, as shown in Figure 1.

In the initial stage, features are projected via three 1×1 convolutions to generate a subset of features represented as 3×N-dimensional feature maps. In the subsequent stage, the feature subsets obtained from the first stage are sent into two independent branches.

The upper branch follows a convolutional path with a kernel size of k, collecting information from a local receptive field. The features are transformed into k² feature maps through a fully connected layer. These generated features are further processed to produce new feature maps. The lower branch follows the self-attention mechanism path, where features are divided into N groups, each containing query, key, and value tensors. These tensors are processed through operations such as shifting and summing to generate new feature maps.

Finally, the outputs of the two branches are combined through a weighted summation, with the weights determined by two learnable scalars, as shown in Equation (1):

$$F_{out} = \alpha F_{conv} + \beta F_{att} \tag{1}$$

Here, $F_{out}$ is the final output of the path; $F_{att}$ is the output of the self-attention branch, and $F_{conv}$ is the output of the convolutional attention branch.

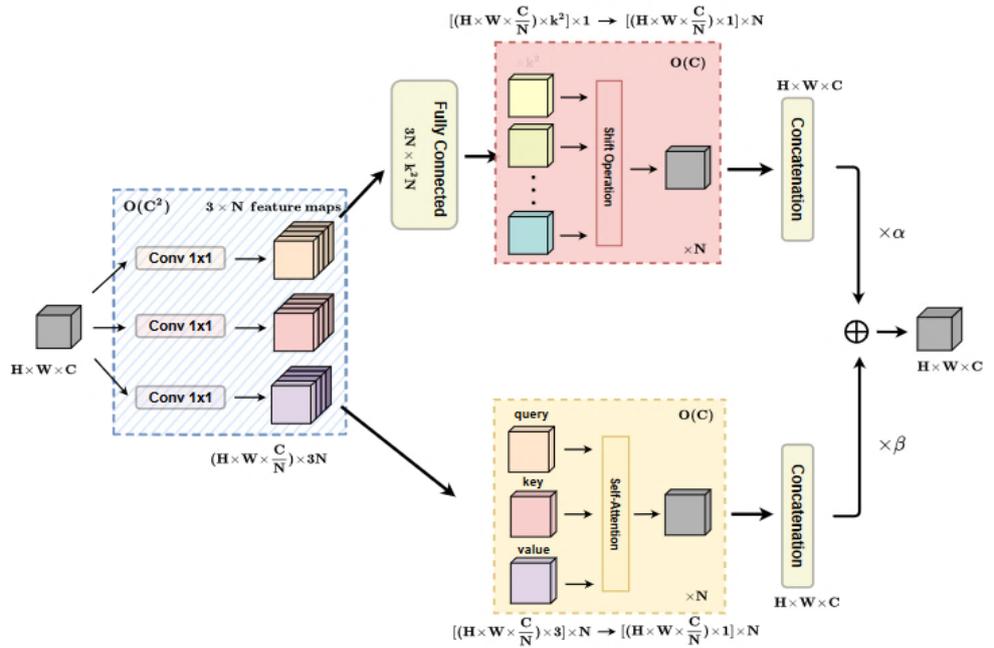

**Figure 1.** ACmix Structure Diagram

## 2.3 BiFormer Sparse Attention Mechanis

The core of the BiFormer sparse attention mechanism is the Bi-level Routing Attention (BRA) mechanism [18]. The structure of BRA is shown in Figure 2. This mechanism divides the input feature map into multiple non-overlapping regions, then applies linear transformations to obtain tensors for query (Q), key (K), and value (V) to calculate attention weights. Fine-grained token-to-token attention is applied within these focused regions, as described in Equations (2) and (3):

$$K^g = gather(K, I^r) \qquad (2)$$

$$V^g = gather(V, I^r) \qquad (3)$$

Here, **gather** is an operation that collects tensors. The resulting Bi-level Routing Attention (BRA) mechanism can be expressed as shown in Equation (4):

$$BRA = Attention(Q, K^g, V^g) + LCE(V) \qquad (4)$$

In Equation (4), LCE(V) represents a local context enhancement term.

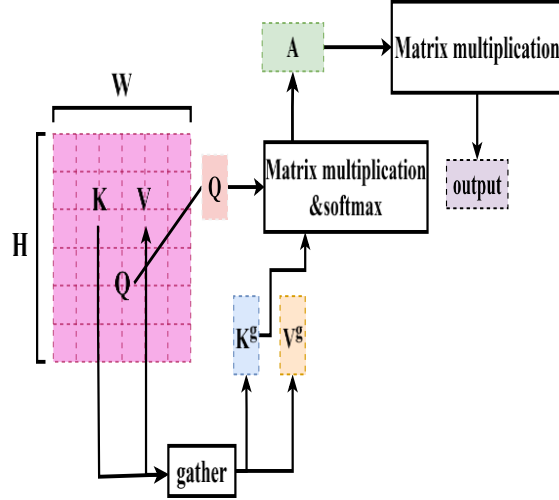

**Figure 2.** BRA Structure Diagram

Based on the BRA module, the BiFormer sparse attention mechanism employs a four-level pyramid structure, as shown on the left side of Figure 3. In each BiFormer block, a 3×3 depthwise convolution is first used to implicitly encode relative positional information. Subsequently, the BRA module and two MLP modules are applied sequentially to model cross-position relationships and per-position embeddings, respectively, as shown on the right side of Figure 3.

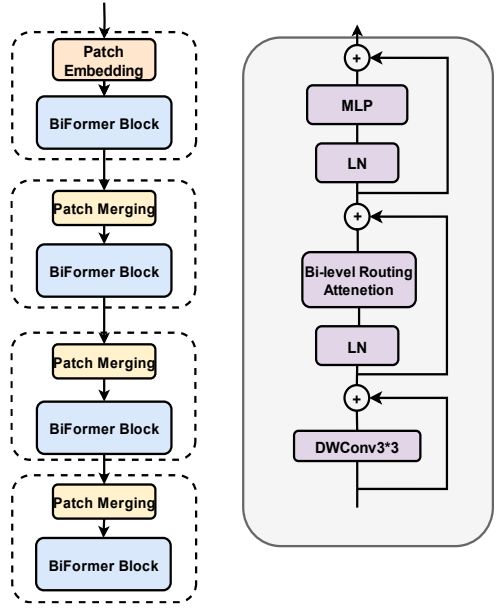

**Figure 3.** Overall Structure Diagram of BiFormer

*2.4 Loss Function*

The original loss function of YOLOv7 is CIoU [19]. This loss function considers the overlapping area, the distance between the center points, and the aspect ratio of the width and height between the predicted bounding box and the ground truth box. The loss function is defined as shown in Equation (5):

$$L_{CIoU} = 1 - IoU + \frac{\rho^2(b, b^{gt})}{c^2} + \alpha v \tag{5}$$

Where IoU is the intersection over union between the predicted bounding box and the ground truth box, cc represents the diagonal distance of the smallest enclosing rectangle of the two boxes, bb is the center point of the ground truth box, $b^{gt}$ is the center point of the predicted box, ϱ\rho denotes the distance between b and $b^{gt}$, $\alpha$ is a weight function, and υ describes the aspect ratio consistency. $\alpha$ and υ are defined as follows:

$$\alpha = \frac{v}{1 - IoU + v} \tag{6}$$

$$v = \frac{4}{\pi^2}\left(\arctan\frac{\omega^{gt}}{h^{gt}} - \arctan\frac{\omega}{h}\right)^2 \tag{7}$$

Where $\omega^{gt}$ and $h^{gt}$ are the width and height of the ground truth box, and $\omega$ and $h$ are the width and height of the predicted box.

## 3. Improved YOLOv7x Detection Method

Due to the significant size differences among anomalies in various types of power equipment defects studied in this paper, and the small size of insulator defects, the original network struggles to accurately extract the location and detail information of these anomalies. To improve detection performance, the

YOLOv7 algorithm was enhanced with the following modifications:

First, an AC-SPPCSPC structure was proposed to replace the original feature extraction structure. By pruning convolutional layers, the number of parameters was reduced while minimizing the filtering of small target information. Additionally, the pooling layer was modified to enhance feature fusion, ultimately improving the network's feature extraction capabilities.

Second, the Biformer attention mechanism was added to the Efficient Aggregation Network (EAGNet) to strengthen the network's focus on key features. Biformer dynamically adjusts the proportion of local and global attention based on task requirements, enhancing the network's flexibility in recognizing different feature images.

Finally, the loss function was replaced to address the issue of mismatched predicted bounding boxes during inference.

The improved network architecture is shown in Figure 4.

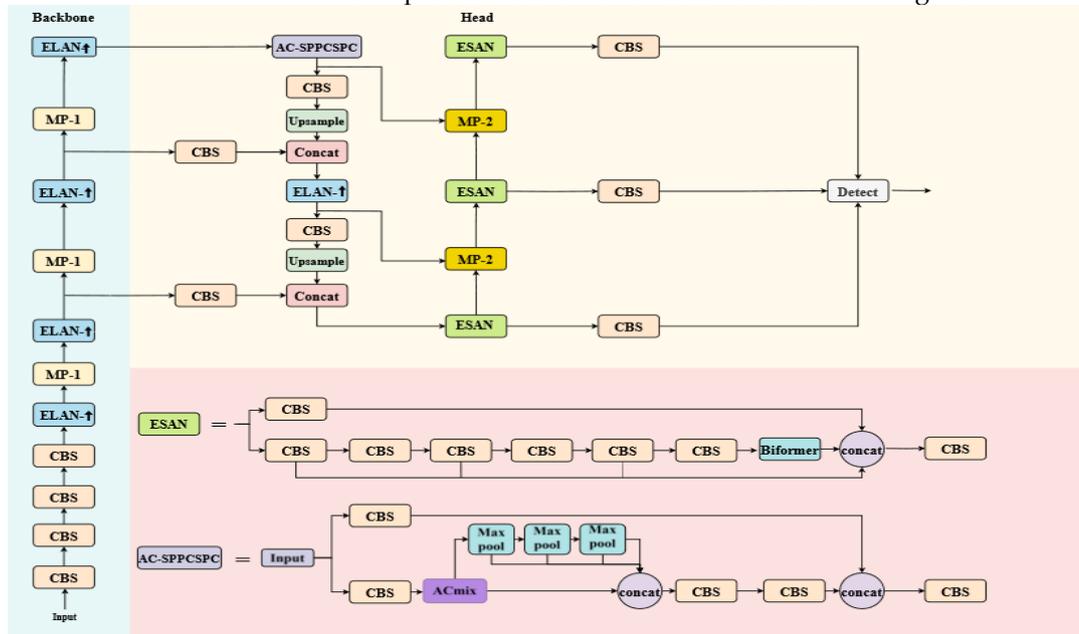

**Figure 4.** Improved YOLOv7 Network Architecture Diagram

*3.1 Improvements to the Feature Extraction Module*

The YOLOv7 series models incorporate the SPPCSPC structure to enhance feature extraction capabilities. This structure processes input feature maps at multiple scales using pooling kernels of various sizes, extracting features from different levels and concatenating them into a composite feature map, thereby improving the model's feature representation. However, SPPCSPC employs a large number of convolutional layers, which increases computational complexity. Additionally, repeated convolutions may blur feature details, resulting in the loss of detailed information and negatively impacting detection performance. To better enhance the feature extraction capability of the network model and address the anomaly detection of power equipment at different scales, this paper introduces improvements to the SPPCSPC feature extraction module.

The specific improvements are as follows:

1) **Introduction of the ACmix Convolutional Mixed Attention Module**

    The ACmix module is introduced before the pooling layer, enhancing the network's flexibility in processing feature maps. This effectively differentiates between various types of defects, improving the model's ability to extract features for multiple types of anomalies. Additionally, by pruning some convolutional layers, the network reduces the filtering of spatial information for defect targets while also lowering computational complexity and parameter size.

2) **Replacing Parallel Pooling Layers with Serial Pooling Layers**

    Compared to the parallel feature extraction method in SPPCSPC, the serial approach offers the same receptive field but achieves faster detection speed with lower computational overhead. Furthermore, by fusing features from the serial pooling layers, the network enhances its ability to represent the location features of defects.

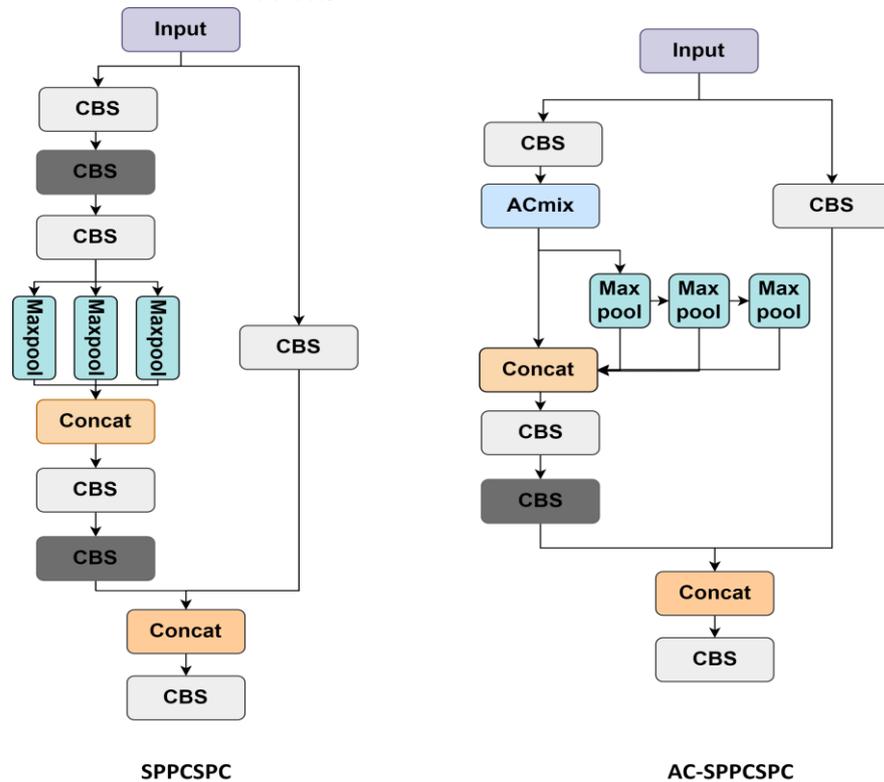

**Figure 5.** SPPCSPC Structure Diagram Before and After Improvement

*3.2 Integration of the BiFormer Attention Mechanism*

The Efficient Aggregation Network (EAGNet) in YOLOv7 uses a multi-branch structure to decompose and combine features, effectively enhancing feature diversity and representation capability. By processing input features through different branches and performing feature aggregation at the end, it ensures effective fusion of multi-layer features.

The BiFormer sparse attention mechanism emphasizes key information and reduces interference from irrelevant data. Through a sparse selection mechanism, it dynamically adjusts the weights of features, highlighting important features while suppressing irrelevant ones. It can also better capture long-range and local detail information, enabling rich feature representation. Furthermore, the sparse attention mechanism can adapt to inputs of different sizes and types, improving the network's flexibility in handling features of varying scales.

In object detection, feature information is crucial for accurately locating and detecting anomalous targets, as these targets may be obscured by larger irrelevant objects or background noise. The BiFormer mechanism enhances the understanding of the scene in an image, improving the accuracy of anomaly detection by strengthening feature representation. Additionally, its strong adaptability allows flexible application to detection scenarios involving anomalies of different sizes and types.

Therefore, this paper embeds the BiFormer sparse attention mechanism into the Efficient Aggregation Network module and names it the Efficient Layer Sparse Attention Aggregation Network (ESAN). The detailed structure is shown in Figure 6.

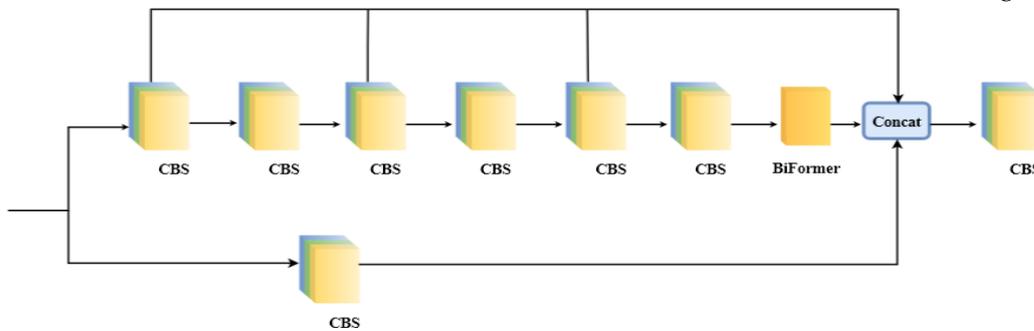

**Figure 6.** ESAN Network Architecture Diagram

*3.3 Improvement of the Loss Function*

The original loss function of YOLOv7 is CIoU. This loss function does not account for image dimensions, making it less effective in optimizing the alignment between predicted and ground truth bounding boxes when images have the same aspect ratio but entirely different width and height values. Additionally, the width and **height** variations in CIoU often exhibit a negative correlation; that is, when one dimension increases, the other decreases. This can lead to mismatched predictions between the bounding box and the actual object, thereby undermining the performance of CIoU in bounding box predictions. Furthermore,

CIoU is insufficiently sensitive to bounding box rotation, which affects detection accuracy in different orientations.

To address these issues, this paper introduces the MPDIoU loss function [20]. MPDIoU provides a more flexible similarity measure for bounding boxes, effectively handling aspect ratio and rotation challenges, thereby improving the accuracy of bounding box matching. Moreover, MPDIoU simultaneously considers the area and shape of the bounding boxes, enhancing robustness in target detection under complex scenarios.

MPDIoU evaluates the distance between the centers of the bounding boxes and uses this distance to adjust the IoU calculation, resulting in a more accurate assessment of bounding box similarity. It incorporates all relevant factors considered by existing loss functions. The computation is as follows:

$$d_1^2 = (x_1^B - x_1^A)^2 + (y_1^B - y_1^A) \tag{8}$$

$$d_2^2 = (x_2^B - x_2^A)^2 + (y_2^B - y_2^A) \tag{9}$$

$$MPDIoU = \frac{A \cap B}{A \cup B} - \frac{d_1^2}{\omega^2 + h^2} - \frac{d_2^2}{\omega^2 + h^2} \tag{10}$$

Here, AA and BB represent the input images, ω\omega and hh denote the width and height of the images, respectively. ($x_1^A$, $y_1^A$) and ($x_2^A$, $y_2^A$) are the coordinates of the top-left and bottom-right points of image A, while ($x_1^B$, $y_1^B$) and ($x_2^B$, $y_2^B$) are the coordinates of the top-left and bottom-right points of image B.

## 4. Experimental Work

*4.1 Experimental Environment Setup*

The hardware environment for this experiment includes a GeForce GTX GPU, an Intel Core i7-12700H CPU, and 16GB of memory. The software environment consists of Windows 11, PyTorch 2.1, Torchvision 0.16, and CUDA 12.2. In the ablation experiments, the stochastic gradient descent (SGD) algorithm was used to train for 100 epochs. The initial learning rate was set to 0.01, with a minimum learning rate of 0.0001, and the batch size was 8. The momentum parameter and weight decay were set to 0.937 and 0.0005, respectively.

*4.2 Dataset Processing*

The study focuses on insulators and dials in power supply networks. The dataset was sourced from field data collected over the years by a power company. A dataset of power equipment defects was constructed using 1,688 images of insulators and 1,270 images of dials, featuring variations in size and background. The LabelImg tool was used to annotate the constructed dataset of power equipment anomalies. Examples of various defect samples are shown in Figure 7.

The annotation labels and their corresponding sample counts are as follows:

Insulator damage (jyz_sh): 1,260 samples

Insulator flashover (jyz_sl): 2,723 samples

Dial blur (bj_mh): 459 samples

Dial damage (bj_ps): 605 samples

Normal dial (bj): 406 samples

Annotations were saved as .xml files. The dataset was split into training, validation, and testing sets in an 8:1:1 ratio.

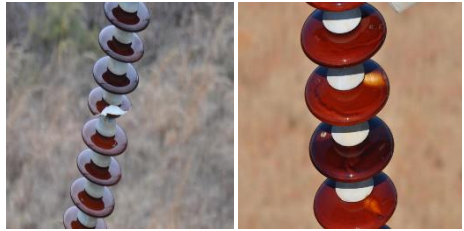

(a）Insulator Damage　　(b）Insulator Flashover

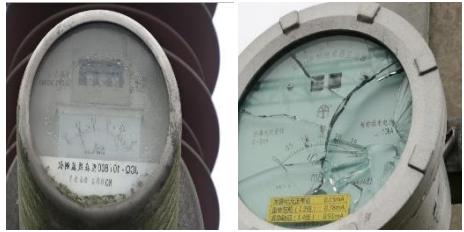

(c）Dial Blur　　(d）Dial Damage

**Figure 7.** Defect Sample Images

## 5. Experimental Results and Comparative Analysis

Authors should discuss the results and how they can be interpreted from the perspective of previous studies and of the working hypotheses. The findings and their implications should be discussed in the broadest context possible. Future research directions may also be highlighted.

*5.1 Ablation Study*

5.1.1 Impact of Different Attention Mechanisms

To investigate the impact of different attention mechanisms, four attention mechanisms—SimAm [23], SE [24], CBAM [27], and BiFormer [18]—were introduced for ablation experiments. The experimental results are shown in Table 1.

From the results in Table 1, it can be observed that among SimAm, SE, CBAM, and BiFormer, the BiFormer attention mechanism introduced in this paper achieved the best performance, with the highest mAP value of 92.6%. This represents an improvement of 1.3% compared to SimAm, 2.7% compared to SE, and 2.1% compared to CBAM. Additionally, the AP values for various anomalies in this dataset were also the highest when using the BiFormer attention mechanism compared to the other three mechanisms. This demonstrates the effectiveness of the BiFormer attention mechanism introduced in this paper.

**Table 1.** Comparison of Accuracy Results for Different Attention Mechanisms

| Attention Mechanism | mAP@0.5/% | AP | | | | |
|---|---|---|---|---|---|---|
| | | bj_mh | bj_ps | bj | jyz_sh | jyz_sl |
| SimAm | 91.3 | 85.8 | 91.4 | 99.6 | 88.2 | 91.5 |
| SE | 89.9 | 84.5 | 91.2 | 99.2 | 86.6 | 87.8 |
| CBAM | 90.5 | 84.6 | 91.7 | 99.5 | 87.8 | 88.9 |
| Biformer | 92.6 | 86.4 | 92.8 | 99.5 | 90.2 | 94.0 |

5.1.2 Impact of the Feature Extraction Module Before and After Improvement

To validate the impact of the improved feature extraction module on target detection accuracy, a heatmap visualization comparison of the module before and after improvement was conducted. The color intensity of the heatmap represents the network's attention to the defect target—the darker the color, the higher the attention to the defect target area. The comparison results are shown in Figure 8.

From the results in Figure 8, it can be observed that compared to the original SPPCSPC module, the heatmap generated using the AC-SPPCSPC module shows colors more concentrated in the defect target regions, with deeper colors. This indicates that the model's attention to defect targets has improved and the perception area has become more accurate with the application of AC-SPPCSPC. This demonstrates that the improved SPPCSPC module can enhance the model's focus on the defect regions in the dataset, reduce interference from complex backgrounds, and improve the model's ability to detect defects.

5.1.3 Ablation Study of Different Improvement Modules

To validate the improvements brought by the added modules to the original network's detection performance, ablation experiments were conducted using the original YOLOv7x without any modifications as a baseline. The methods tested include incorporating the improved AC-SPPCSPC module, embedding the BiFormer attention mechanism, and optimizing the loss function. A "√" indicates the use of a particular module combination. The specific experimental results are shown in Table 2.

After training, the original YOLOv7x achieved a precision of 95.9%, a recall of 95.9%, and an mAP of 89.2%. Comparing the results after replacing the loss function, it was found that substituting the original loss function with MPDIoU increased the mAP by 1%, precision by 1.1%, and recall by 1%. Further comparison of performance metrics after separately adding AC-SPPCSPC and the BiFormer attention mechanism to YOLOv7x revealed mAP improvements of 2.7% and 3.2%, respectively.

Ultimately, the improved YOLOv7x network model proposed in this paper achieved a 4.3% increase in mAP compared to the original model, along with improved precision and recall. Moreover, the parameter count decreased from 70.8M to 67.3M with the inclusion of the proposed modules. Therefore, the final ablation experiment results indicate that the proposed improvement modules positively impact the detection accuracy of the original network, with the best results achieved when all three modules are applied simultaneously.

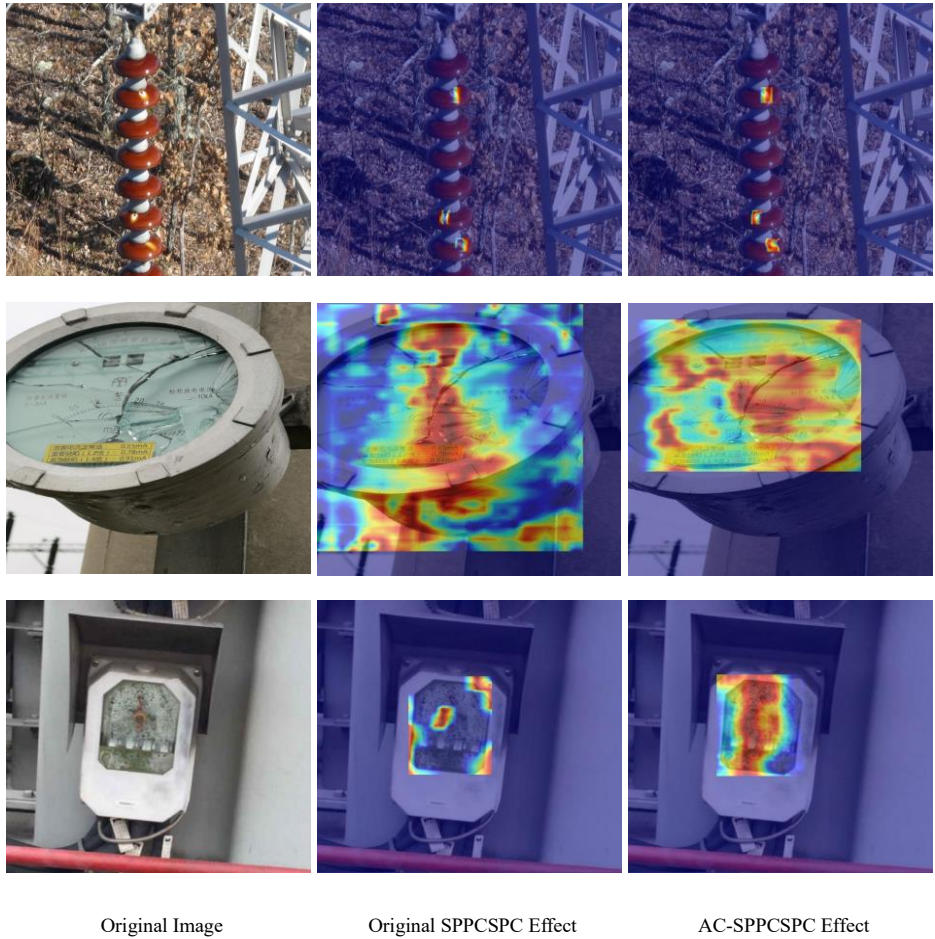

Original Image    Original SPPCSPC Effect    AC-SPPCSPC Effect

**Figure 8.** Heatmap Comparison Before and After SPPCSPC Improvement

**Table 2.** Comparison of Ablation Experiment Results for the Improved YOLOv7x Model

| Group | AC-SPPCSPC | MPDIoU | Biformer | P/% | R/% | mAP@0.5/% | Number of Parameters / M |
|---|---|---|---|---|---|---|---|
| A | | | | 95.9 | 95 | 89.2 | 70.8 |
| B | √ | | | 96.6 | 96.3 | 91.9 | 69.3 |
| C | | √ | | 96.8 | 96 | 90.2 | 70.8 |
| D | | | √ | 95.6 | 97 | 92.4 | 68.7 |
| E | √ | √ | | 96.2 | 97 | 92.6 | 69.3 |
| F | √ | | √ | 97 | 97 | 92.8 | 67.3 |
| G | | √ | √ | 96.5 | 97 | 92.7 | 68.7 |
| H | √ | √ | √ | 97.1 | 97 | 93.5 | 67.3 |

*5.2 Comparative Experiments*

5.2.1 Comparison with Mainstream Models

To verify the effectiveness of the improved model, it was compared with other commonly used object detection models, such as SSD [2], YOLOv7 [21], DETR [25], TPH-YOLOv5 [26], and YOLOv8, under identical conditions. All models were trained on a unified experimental platform and used the same training and testing datasets.

From the results in Table 3, the proposed algorithm outperformed other mainstream algorithms in terms of mAP values for defects such as insulator damage, flashover, and dial blur and damage. The improved algorithm achieved an mAP of 93.5%, which is a 4.3% improvement compared to the original algorithm and an increase over other commonly used object detection algorithms.

A detailed comparison with the original algorithm revealed the following improvements in AP values for specific defect types:
- Dial blur (bj_mh): 3.9% improvement
- Dial damage (bj_ps): 2.2% improvement
- Insulator damage (jyz_sh): 6.9% improvement
- Insulator flashover (jyz_sl): 8.7% improvement

The experimental results strongly demonstrate the effectiveness of the improved algorithm proposed in this paper.

**Table 3.** Model Comparison Experiment Results

| Algorithm | mAP@0.5/% | AP | | | | |
|---|---|---|---|---|---|---|
| | | bj_mh | bj_ps | bj | jyz_sh | jyz_sl |
| SSD | 79.1 | 76.9 | 80.8 | 97 | 71.2 | 69.6 |
| Yolov5 | 83.4 | 78.9 | 85.8 | 98.4 | 77.4 | 76.2 |
| Yolov7 | 89.2 | 83.9 | 90.7 | 99.5 | 85.4 | 86.5 |
| Yolov8 | 90.3 | 84.8 | 90.1 | 99 | 87.4 | 90.2 |
| DETR | 90.2 | 85.4 | 90.6 | 99.5 | 88.1 | 87.3 |
| Ours | 93.5 | 87.8 | 92.9 | 99.5 | 92.3 | 95.2 |

5.2.2 Comparative Analysis with the Original Model

To further validate the effectiveness of the improvements in the proposed model, a visual comparison of the detection results before and after the improvements was conducted, as shown in Figure 8.

From the results of the first three groups, it can be observed that the original model exhibits false detections in complex backgrounds with densely packed insulator defects. In the first and second groups, the original model incorrectly detected non-damaged insulators as damaged. In the third group, it falsely identified non-flashover cases as insulator flashovers. In contrast, the proposed model accurately detected these cases, with significantly better confidence scores.

In the fourth group, the original model mistakenly identified non-blurred dials as blurred, whereas the proposed model did not exhibit this issue.

These results demonstrate that the improvements in the SPPCSPC structure have enabled the network to more effectively extract target features. Additionally, embedding the BiFormer attention mechanism in the Efficient Aggregation Network has enhanced the distinguishability of features for different targets. Replacing the loss function has effectively reduced the false detection rate, thereby improving the accuracy of target detection.

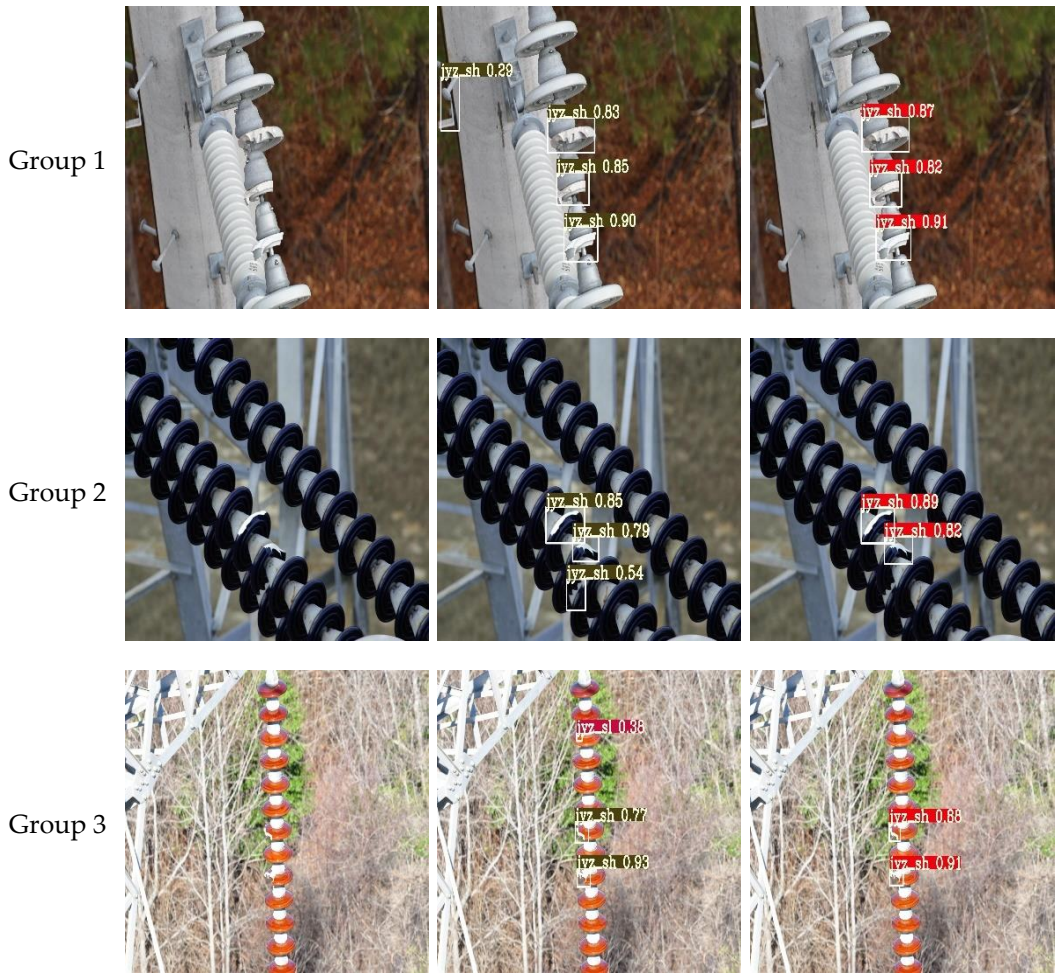

Group 1

Group 2

Group 3

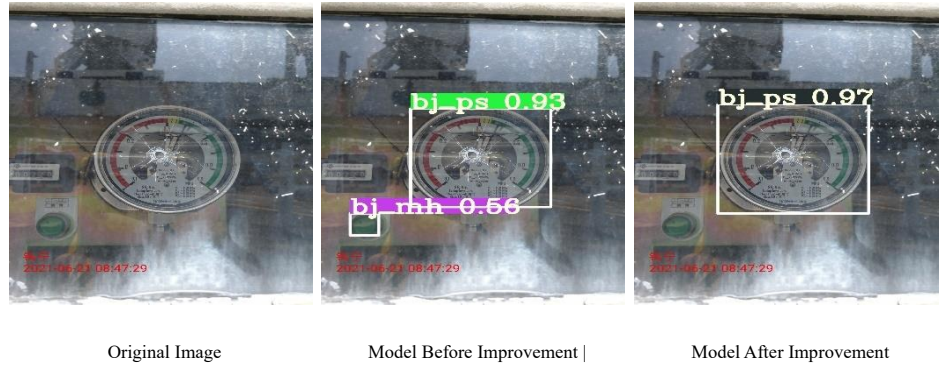

Group 4 — Original Image | Model Before Improvement | Model After Improvement

**Figure 9.** Visualization Comparison of Defect Detection Results

## 5. Conclusions

This paper proposes an improved YOLOv7x network model for anomaly detection in power system equipment. The main improvements include the following:

1) **Improvement of the SPPCSPC Module**:
   A new SPPCSPC structure was proposed for the backbone network. The number of convolutional layers before the pooling layers was reduced to decrease the parameter count, and the ACmix convolutional mixed attention mechanism module was introduced to enhance the feature extraction capability for power equipment images obtained during intelligent inspections. Additionally, the parallel pooling layers in SPPCSPC were replaced with serial pooling layers, improving the model's ability to focus on defect targets.

2) **Integration of Sparse Attention Mechanism**:
   A sparse attention mechanism was incorporated into the Efficient Aggregation Network (EAGNet), resulting in the proposed ESAN network structure. This enhancement improved the network's flexibility in identifying defects in different types of equipment.

3) **Optimization of the Loss Function**:
   The original loss function was replaced with the MPDIoU function, addressing the issue of mismatched target prediction bounding boxes.

Experimental results validated the effectiveness of the proposed algorithm, achieving an mAP of 93.5%, a Precision of 97.1%, and a Recall of 97%. The improved algorithm significantly reduced occurrences of missed and false detections.

Future work will focus on further optimizing the algorithm for eventual application in power systems.

**Funding:** This research was funded by the State Grid Sichuan Electric Power Company Science and Technology Program, grant number 521997230014, and the funding of Southwest Jiaotong University's first batch of English-taught quality courses for international students in China, grant number LHJP[2023]07.

**Institutional Review Board Statement:** Not applicable.

**Informed Consent Statement:** Not applicable.

**Data Availability Statement:** Substation instrumentation image data is not available due to privacy.

**Conflicts of Interest:** The authors declare no conflicts of interest.